\def\year{2022}\relax
\documentclass[letterpaper]{article} 
\usepackage{aaai22}  
\usepackage{times}  
\usepackage{helvet}  
\usepackage{courier}  
\usepackage[hyphens]{url}  
\usepackage{graphicx} 
\usepackage{natbib}  
\usepackage{caption} 
\DeclareCaptionStyle{ruled}{labelfont=normalfont,labelsep=colon,strut=off} 
\usepackage{algorithm}
\usepackage{algorithmic}
\usepackage{dsfont}
\usepackage{booktabs}
\usepackage{multirow}
\usepackage{makecell}
\usepackage{xspace}
\usepackage{xcolor}
\usepackage{amsmath}
\usepackage{amssymb}
\usepackage{enumitem}
\usepackage{subfigure}

\usepackage{newfloat}
\usepackage{listings}
\floatstyle{ruled}
\newfloat{listing}{tb}{lst}{}
\floatname{listing}{Listing}
\newcommand{\myparagraph}[1]{\vspace{3.0pt}\noindent{\bf #1}}

\makeatletter
\DeclareRobustCommand\onedot{\futurelet\@let@token\@onedot}
\def\@onedot{\ifx\@let@token.\else.\null\fi\xspace}

\def\eg{\emph{e.g}\onedot} 
\def\ie{\emph{i.e}\onedot}

\def\wrt{w.r.t\onedot}

\renewcommand*{\@fnsymbol}[1]{\ensuremath{\ifcase#1\or \dagger\or *\or \ddagger\or
   \mathsection\or \mathparagraph\or \|\or **\or \dagger\dagger
   \or \ddagger\ddagger \else\@ctrerr\fi}}

\makeatother

\title{Keypoint Message Passing for Video-based Person Re-Identification}
\author{
    Di Chen\thanks{Equal contribution.},\textsuperscript{\rm 1,}\textsuperscript{\rm 3}
    Andreas D\"oring$^\dagger$,\textsuperscript{\rm 2}
    Shanshan Zhang\thanks{Corresponding author.},\textsuperscript{\rm 1}
    Jian Yang$^*$,\textsuperscript{\rm 1}
    Juergen Gall,\textsuperscript{\rm 2}
    Bernt Schiele\textsuperscript{\rm 3}
}
\affiliations{
    \textsuperscript{\rm 1} PCA Lab, Key Lab of Intelligent Perception and Systems for High-Dimensional Information of Ministry of Education, Jiangsu Key Lab of Image and Video Understanding for Social Security, \\Nanjing University of Science and Technology\\
    \textsuperscript{\rm 2} University of Bonn\\
    \textsuperscript{\rm 3} Max Planck Institute for Informatics\\
    \texttt{\{dichen,shanshan.zhang,csjyang\}@njust.edu.cn, \{doering,gall\}@iai.uni-bonn.de, \{dichen,schiele\}@mpi-inf.mpg.de}
}

\begin{document}
\maketitle
\thispagestyle{plain} 
\pagestyle{plain} 

\begin{abstract}
Video-based person re-identification~(re-ID) is an important technique in visual surveillance systems which aims to match video snippets of people captured by different cameras.
Existing methods are mostly based on convolutional neural networks~(CNNs), whose building blocks either process local neighbor pixels at a time, or, when 3D convolutions are used to model temporal information, suffer from the misalignment problem caused by person movement. 
In this paper, we propose to overcome the limitations of normal convolutions with a human-oriented graph method.
Specifically, features located at person joint keypoints are extracted and connected as a spatial-temporal graph.
These keypoint features are then updated by message passing  from their connected nodes with a graph convolutional network~(GCN). 
During training, the GCN can be attached to any CNN-based person re-ID model to assist representation learning on feature maps, whilst it can be dropped after training for better inference speed.
Our method brings significant improvements over the CNN-based baseline model on the MARS dataset with generated person keypoints and a newly annotated dataset: PoseTrackReID.
It also defines a new state-of-the-art method in terms of top-1 accuracy and mean average precision in comparison to prior works.\footnote{The new dataset and code will be released at \url{https://github.com/DeanChan/KeypointMessagePassing}.} 
\end{abstract}

\section{Introduction}
\begin{figure}[t] 
    \centering
    \includegraphics[width=0.5\textwidth]{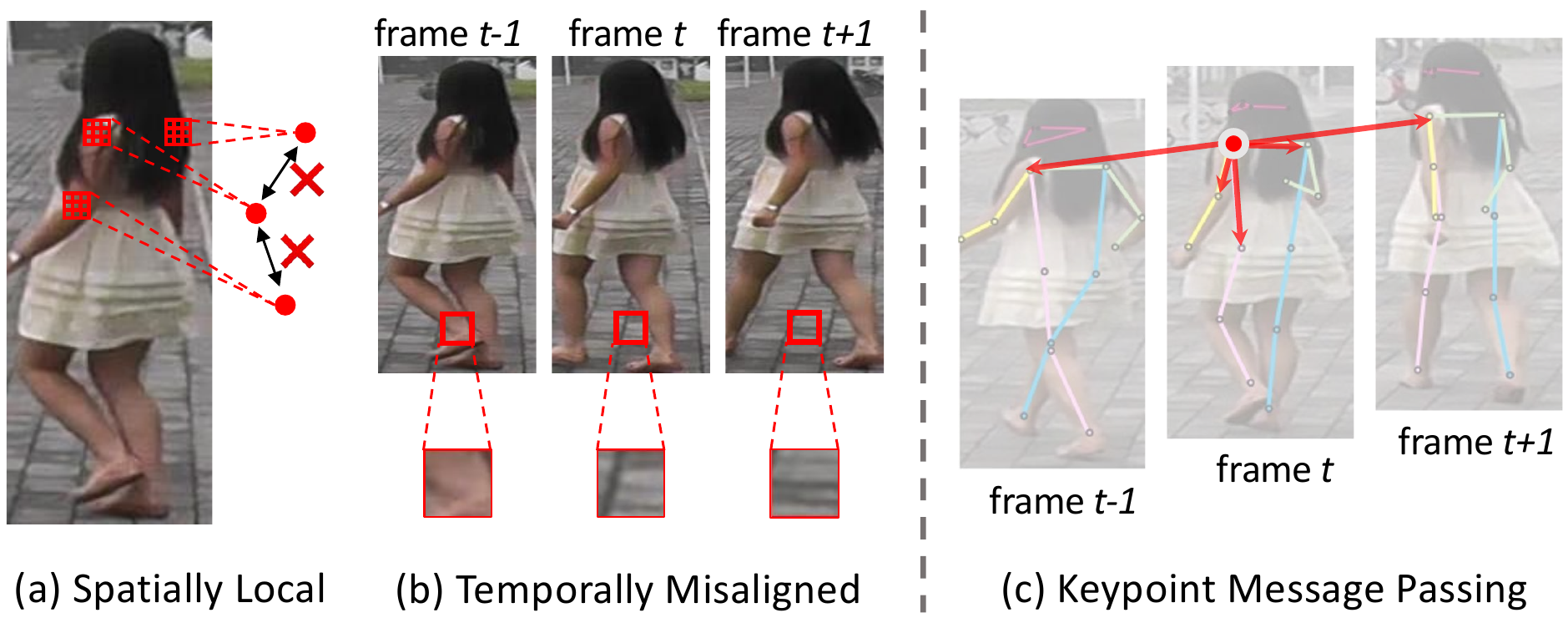}
    \caption{Problems using 2D/3D convolution: (a) spatially local and (b) temporal misalignment;
    We propose (c) \textbf{keypoint message passing}, where keypoint features on frame $t$ (marked with white circle) are updated with information from connected nodes (red arrow) by \textit{graph convolution}, which is not bounded by the fixed shape and location of normal convolutions. The complete spatial-temporal graph structure is shown in Fig.~\ref{fig:skelenton}.
    }
    \label{fig:teaser}
    \vspace{-3pt}
\end{figure}

Visual surveillance systems play a vital role in modern society to ensure public safety.
The massive data collected by the systems raises the need for automatic visual understanding techniques such as person re-identification~(re-ID). 
Person re-ID aims to associate the same person across cameras with non-overlapping views, which is usually achieved by calculating the similarities between the representations of images.
Compared to images, video sequences provide much richer information which is beneficial to address visual ambiguities.
Therefore, video-based person re-ID emerged recently as a parallel research field to image-based person re-ID.

The central problem of learning discriminative video representations is how to exploit both spatial and temporal information.
Most existing solutions propose to capture spatial and temporal information separately, \ie using a 2D convolutional neural network~(CNN) for spatial representation learning, while handling temporal information by aggregating the high-level outputs of CNNs with pooling~\cite{zheng2016mars}, recurrent neural networks~\cite{chung2017two,mclaughlin2016recurrent,chen2018video,xu2017jointly,zhou2017see} or temporal attention~\cite{fu2019sta,zhou2017see,xu2017jointly,liu2017quality,li2018diversity}.
Other works~\cite{liu2019spatial,liu2019spatially,li2019multi} learn concurrent spatial-temporal representations with 3D convolutions.
The core operation of both types of methods is convolution, which only processes information at a small local range, especially when it is located at shallow hierarchies of a network.
Meanwhile, 3D convolution layers also have the misalignment problem, \ie the same position in adjacent frames may belong to different body parts due to person movement, such that the appearance representations learned by 3D convolution are polluted with the wrong body part or background.
The spatially local and temporally misalignment problems are shown in Fig.~\ref{fig:teaser} (a) and (b).

In this paper, we propose to overcome the limitations of convolutions in video-based person re-ID with the help of person pose estimation and graph convolution.
Since human body keypoints are representative for distinct shapes of different people, we construct the graph based on keypoints.
Specifically, the features located at person joint keypoints are cropped from the CNN feature maps, which are then used as the node features for constructing a spatial-temporal graph.
The graph-structured data is then processed with graph convolutions, which is not bounded by small-scale, rectangular-shaped kernels of normal convolution.
A brief illustration is shown in Fig.~\ref{fig:teaser} (c).
We can see that features located at, \eg one's left shoulder, interact with features from the right shoulder, left elbow and left hip simultaneously, which is otherwise hard to reach for a single convolutional layer due to the large spatial distances.
Meanwhile, the left shoulder features at video frame $t$ also receive information from the same location at adjacent frames without misalignment problems.
The same process happens to all other keypoints on the human body, enriching the local keypoint features with non-local, temporal-aligned and human-oriented information.
Since the core idea is based on the message passing mechanism~\cite{gilmer2017neural} of graph convolutions, we name our method \textit{keypoint message passing}.

\begin{figure*}[t]
    \centering
    \includegraphics[width=0.95\linewidth]{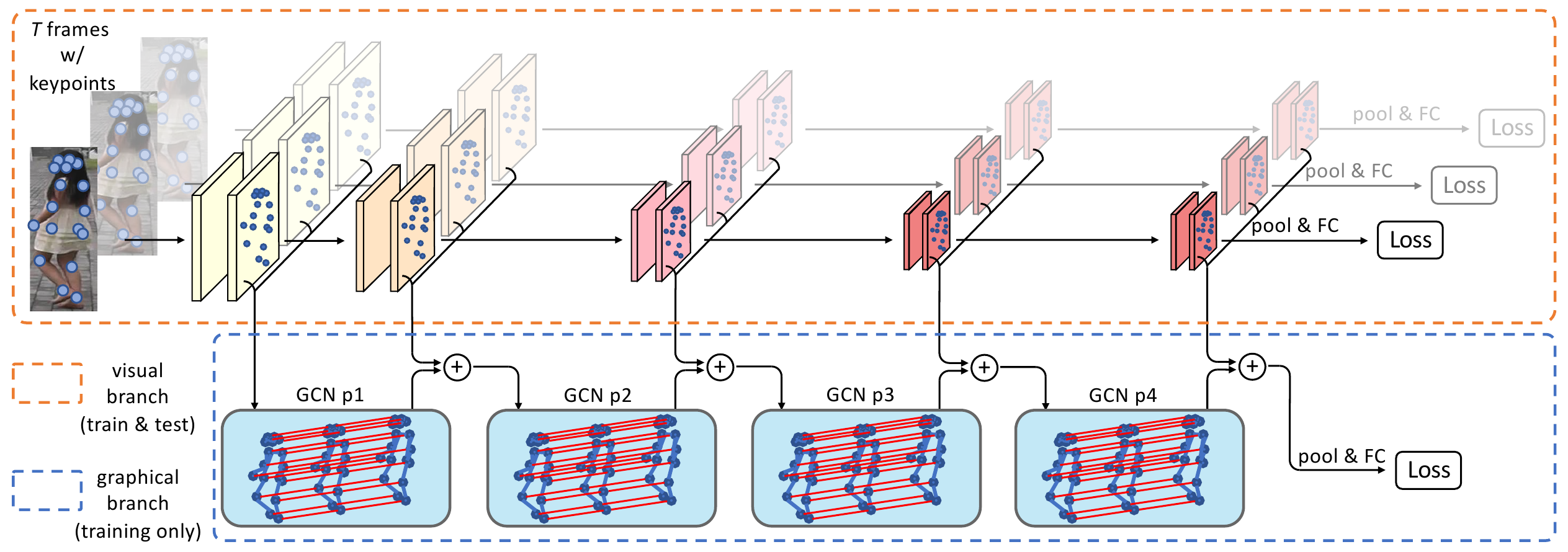}
    \caption{\textbf{Overall pipeline for our method.} 
    The \textit{visual branch} is a base CNN which is a typical 5-stage model and takes in a video with $T$ frames as input.
    The \textit{graphical branch} is a GCN divided into 4 parts. 
    Given the keypoint locations (either annotated or generated with a pose estimation model), we extract features according to the locations on the first CNN stage and use them as the inputs to the GCN.
    At the end of each stage, keypoint features are fused with the intermediate representations from GCN with an element-wise sum. The fused features serve as the new inputs to the subsequent GCN layers. The detailed graph topology is shown in Fig.~\ref{fig:skelenton}. Both the CNN and GCN are supervised with cross-entropy loss during training. Note, that the GCN branch including keypoint estimation can be dropped during inference.}
    \label{fig:overview}
    \vspace{-3pt}
\end{figure*}

When we attach a graph convolutional network~(GCN) to a CNN, it comes with significant computational cost and extra memory consumption, especially when the graph scale is large and the GCN is deep.
To this end, we propose a flexible design which enables graph convolutions during training but does not require the graph during inference.
An illustration is shown in Fig.~\ref{fig:overview}.
The GCN serves as a parallel branch along side the CNN, which functions as a training guide.
Supervision signals with spatial-temporal information flows back from the GCN to the CNN by back-propagation.
Therefore, each keypoint location on the CNN feature map receives feedback from a more diverse set of spatial-temporal locations, especially for the shallow layers which have small receptive fields.
Once the model is trained, the whole GCN branch, keypoints and graphs can be dropped, leaving no extra computational burden other than the CNN branch.
Moreover, the choice of the CNN is not limited, \ie our design is generalizable to any CNN-based re-ID model.
We name our method `KMPNet'.

In summary, our contribution is three-fold:
\begin{itemize}
    \vspace{-3pt}
    \item With the help of spatial-temporal guidance provided by person joint keypoints and graph convolutions, we propose a general method to assist CNN-based re-ID model training, which overcomes the limitation of normal convolutions without extra computational burden during inference. 
    \vspace{-3pt}
    \item We present PoseTrackReID, a new dataset for video-based re-ID featuring both person ID and keypoint annotations.  
    \vspace{-3pt}
    \item Extensive experiments demonstrate that our model significantly improves the baseline, achieving results on par with or better than the current state-of-the-art models.
\end{itemize}



\section{Related Work}
\myparagraph{Image-based Person Re-Identification.} 
Image-based person re-ID models usually serve as good baselines for video-based re-ID methods.
Early re-ID models mainly focus on designing discriminative hand-crafted features~\cite{Wang2007,Farenzena2010,Zhao2013,Liao_2015_CVPR} and distance metrics~\cite{Kostinger2012,Li2015,Zhang_2016_CVPR}.
Nowadays, designing re-ID models based on CNNs has become main stream.
These methods typically formulate re-ID as a classification~\cite{xiao2016learning,Zheng2016,Fan2018SphereReIDDH,Xiang2018HomocentricHF} or ranking~\cite{Yi2014,Li2014,Ahmed2015,Varior2016,Liu2017,xu2018attention} problem at training time, and use the optimized backbone network as a feature extractor during inference.
Instead of extracting global features with global average pooling, recent methods~\cite{sun2018beyond,wei2017glad,zhao2017deeply,yao2019deep} propose to divide the final CNN feature maps into several parts and use average pooling separately.
For example, PCB~\cite{sun2018beyond} partitions the feature maps into horizontal stripes and then concatenates the pooled stripe features to generate the final features, which contain richer spatial information and thus achieve better performance than the simple global features.
In our ablation studies, we also choose PCB as the base CNN of the visual branch, whereas it is worth to notice that any CNN based model is a candidate.

\myparagraph{Video-based Person Re-Identification.} 
The most direct way for video-based re-ID is to lift image-based re-ID methods by aggregating multi-frame features via different operations, such as mean/max pooling~\cite{zheng2016mars}, recurrent neural networks~(RNNs)~\cite{chung2017two,chen2018video,xu2017jointly,zhou2017see} and temporal attention~\cite{fu2019sta,zhou2017see,xu2017jointly,liu2017quality,li2018diversity}. 
Another strategy is to capture the spatial and temporal information simultaneously by 3D convolution~\cite{liu2019spatial,liu2019spatially,li2019multi}.  
Despite their favorable performance, 3D convolutions usually require more computational and memory resources, which limits their potential for real-world applications.
The graph convolution in our method shares similar concept with 3D convolution on concurrent spatial-temporal information modeling.
However, our method differs from 3D CNNs in that the input data has a non-local structure,
whereas the input to 3D convolutions must be a rectangular-shaped local range of pixels.
Besides, our method does not suffer from the temporal misalignment problem as 3D convolutions do, since the cross-frame features are always extracted at the same location on the human body.  
Additionally, all the graph convolution operations are only performed at training time, thus no extra computational costs are required during inference.  

\myparagraph{Pose-assisted Person Re-Identification.} 
Benefiting from recent advances on pose estimation~\cite{rafi2020selfsupervised,openpose,xiao2018simple,sun2019hrnet}, person keypoints have been utilized to facilitate person re-ID models.
Some works~\cite{ge2018fd,liu2018pose} focus on generating person images with keypoints which are later used as extra training data.
Others propose to use person keypoints as a spatial cue for aligning body parts~\cite{suh2018part,su2017pose,wu2020adaptive} or highlighting non-occluded regions~\cite{miao2019pose}.
For instance, \citet{su2017pose} crop out person parts from the input images according to provided keypoints and re-assemble them into a pose-normalized synthetic image.
\citet{miao2019pose} use the keypoint heatmaps as spatial attention which is then multiplied with the feature maps element-wise before average pooling.
In our method, person keypoints also play a key role.
However, they are introduced with a different motivation, \ie refining CNN features by capturing non-local human-oriented information within video frames as well as temporal information between frames.

\myparagraph{Graph Convolutional Networks.} 
The growing need for processing non-Euclidean data has motivated research on graph convolutional networks~\cite{kipf2016semi,chen2018fastgcn,hamilton2017inductive,huang2018adaptive,li2020deepergcn}. 
Some computer vision researchers also take advantage of GCNs on tasks such as action recognition~\cite{yan2018spatial}, video classification~\cite{wang2018videos} and gait recognition~\cite{li2020model}. 
Meanwhile, some person re-ID works~\cite{wu2020adaptive,shen2018person,yang2020spatial,yan2020learning} also exploit GCNs for unstructured relationship learning.
\citet{shen2018person} builds a graph based on probe-gallery image pairs and utilize a GCN for better similarity estimation.
\citet{shen2018person},\citet{yang2020spatial} and \citet{yan2020learning} propose to model the relationship of intra-frame spatial parts and inter-frame temporal cues for a video.
Our method differs from these works in the following three aspects.
Firstly, instead of using horizontal parts as the graph node, we use person keypoints which provide better localization on the human body, and thus avoid the misalignment problem naturally.
Secondly, the graph convolution for spatial-temporal refinement is applied to all levels of CNN features, whereas \citet{shen2018person},\citet{yang2020spatial} and \citet{yan2020learning} only use a GCN for high-level features.
Thirdly, the graphical branch with graph convolutions is only used to assist CNN feature training. 
While during inference, it can be dropped entirely to save computation and memory resources.


\section{Methodology}

\subsection{The Framework}
A brief overview of our model is shown in Fig.~\ref{fig:overview}.
Our model is mainly composed of two branches, \ie the \textit{visual} branch and the \textit{graphical} branch. 
The visual branch is a base CNN model, as a canonical choice for person re-identification.
The input to the visual branch is a video with $T$ frames, while each frame is processed as an individual image.
The graphical branch is a GCN model, which is manually partitioned into 4 stages to match the hierarchies of the base CNN.
We extract the person keypoint features with an RoIAlign~\cite{he2017mask} operation from the last layer of each CNN stage, which are further constructed into a spatial-temporal graph and serve as input to the corresponding GCN stages for long-range interaction. 
The keypoint locations are manually annotated or generated with an off-the-shelf pose estimation model.
Specifically, keypoint features from CNN stage 1 are used as the initial input to GCN part 1.
After passing through several graph convolution layers, the outputs of GCN part 1 are then fused with keypoint features from CNN stage 2 with an element-wise sum.
The ``convolution-fusion'' paradigm is repeated until the end of both branches. 
During training, both of the branches are supervised with cross-entropy losses. 
Gradients from the graphical branch flow back to the visual branch, which enhances the features located at person joints with long-range information.
During inference, we only use the enhanced visual features produced by the CNN for person matching.
At this time, the whole GCN branch can be dropped to reduce the computational burden.

\subsection{Graph Formulation}
\begin{figure}[t]
	  \centering
      \subfigure{
            \centering
	        \includegraphics[width=0.12\textwidth]{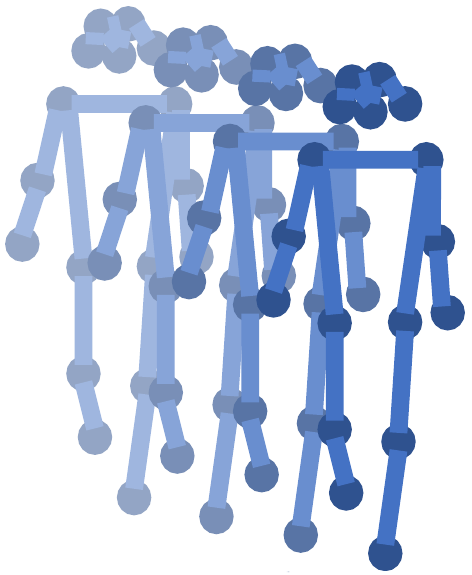}
            \label{fig:skelenton_spatial}}
      \subfigure{
            \centering
	        \includegraphics[width=0.12\textwidth]{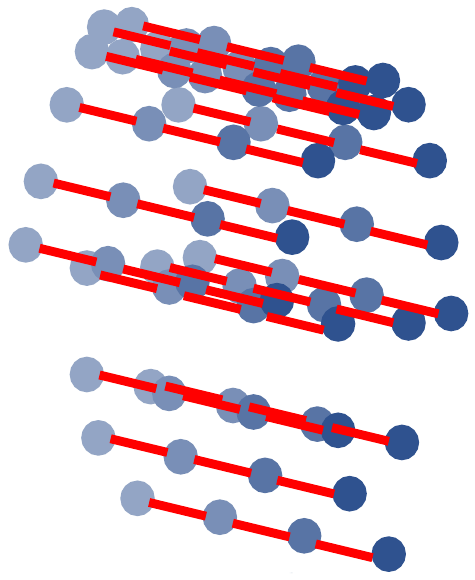}
            \label{fig:skelenton_temporal}}
      \subfigure{
            \centering
	        \includegraphics[width=0.12\textwidth]{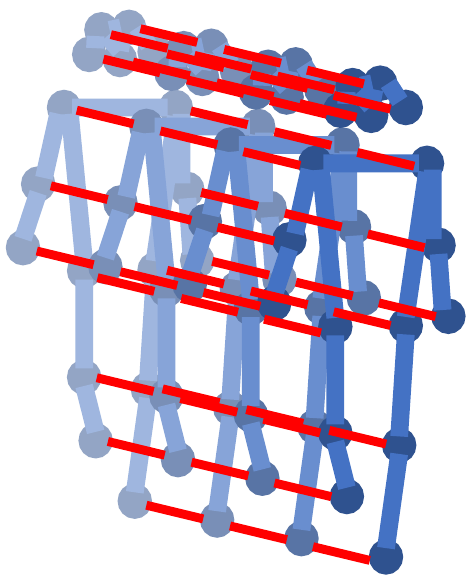}
            \label{fig:skelenton_all}}
	  \caption{\textbf{Spatial-temporal graph for a sequence of person joints.} Figures from left to right denote spatial edges $E_S$ (blue), temporal edges $E_T$ (red) and all edges $E$ respectively. We use $E$ in our KMP method unless specified otherwise.}
	  \label{fig:skelenton}
	  \vspace{-3pt}
\end{figure}

Following \cite{yan2018spatial} and \cite{li2020model}, we represent a sequence of person joints with a spatial-temporal graph $G = (V, E)$.
The node set $V = \{v_{ti} | t=1,\ldots,T, i=1,\ldots,N\}$ includes all the joints in this sequence, where $T$ is the number of frames and $N$ is the number of keypoints in each frame.
The edge set $E$ is composed of two subsets, namely \textit{spatial} set $E_S$ and \textit{temporal} set $E_T$.
The spatial edges, denoted as $E_S = \{ (v_{ti},v_{tj}) | (i, j) \in H, t=1,\ldots,T \}$, is a direct representation of natural human topology $H$ in each frame, as is shown by the blue lines in Fig.~\ref{fig:skelenton}.
The temporal set $E_T = \{(v_{ti},v_{(t+1)i}), | t=1,\ldots,T, i=1,\ldots,N \}$ consists of connections of the same joints between frame $t$ and $t+1$.
It is illustrated as the red lines in Fig.~\ref{fig:skelenton}.

There are different options for features of each graph node. 
In \cite{yan2018spatial} and \cite{li2020model}, the keypoint coordinates are used as node features in order to capture the action or gait information.
Differently, we use visual features cropped from CNN feature maps as graph node feature, since our motivation is to capture the non-local, temporal-aligned relationships between different locations, rather than modeling movement information. 
Specifically, the feature for node $v$ at layer $l$ is represented as $\mathbf{h}_{v}^{(l)}$.
It has three variants depending on which layer it is processed.
For the first layer of GCN p1, $\mathbf{h}_{v}^{(l)}$ is the plain keypoint feature cropped from the feature maps of CNN stage 1.
For the hidden layers inside each GCN part, $\mathbf{h}_{v}^{(l)}$ is the latent outputs from layer $l-1$.
The input features for the first layers of GCN p2 to GCN p4 are linear combinations of the latent outputs and CNN features:
\begin{align}\label{eqn:fusion}
    \mathbf{h}_{v}^{(l)} \leftarrow \mathbf{W}_{down}^T(\mathbf{W}_{up}^T\mathbf{h}_{v}^{(l)} + f(i, j)),
\end{align}
where $f$ denotes the CNN feature map and $i, j$ are the spatial coordinates of node $v$; 
$\mathbf{W}_{up}$ and $\mathbf{W}_{down}$ are weights of fully-connected layers to match the dimensions of $\mathbf{h}_{v}^{(l)}$ and $f(i, j)$.

\subsection{Keypoint Message Passing}\label{sec:KMP}
Once the graph topology and node features are defined, graph convolutions could be applied to update the node features.
We adopt the improved version of graph convolution block from \cite{li2020deepergcn}, which takes advantage of generalized message aggregation, modified skip connections and a novel normalization method.
The block consists of a series of operations, including normalization, non-linear activation, dropout, graph convolution and residual addition.
For simplicity, we only introduce the graph convolution operation since the message passing between different nodes only happens at this step.

For GCN layer $l$, the graph convolution is mainly composed of three actions, namely \textit{message construction} $\boldsymbol{\rho}$, \textit{message aggregation} $\boldsymbol{\zeta}$ and \textit{vertex update} $\boldsymbol{\phi}$.
They are defined as follows respectively:
\begin{align}
    \mathbf{m}_{vu}^{(l)} &= \boldsymbol{\rho}^{(l)}(\mathbf{h}_{v}^{(l)}, \mathbf{h}_{u}^{(l)}), u \in \mathcal{N}(v) \\
    \mathbf{m}_{v}^{(l)} &= \boldsymbol{\zeta}^{(l)}(\{ \mathbf{m}_{vu}^{(l)} |  u \in \mathcal{N}(v) \}) \\
    \mathbf{h}_{v}^{(l+1)} &= \boldsymbol{\phi}^{(l)}(\mathbf{h}_{v}^{(l)}, \mathbf{m}_{v}^{(l)})
\end{align}
where $\mathcal{N}(v)$ denotes the neighbour nodes of vertex $v$;
$\mathbf{m}_{vu}^{(l)}$ indicates the message passed from node $u$ to $v$; 
$\mathbf{m}_{v}^{(l)}$ is the aggregated messages for node $v$.
Since we do not use edge features here, the message construction function $\boldsymbol{\rho}^{(l)}(\cdot)$ is simply defined as:
\begin{align}\label{eqn:rho}
    \mathbf{m}_{vu}^{(l)} = \boldsymbol{\rho}^{(l)}(\mathbf{h}_{v}^{(l)}, \mathbf{h}_{u}^{(l)}) = \text{ReLU}(\mathbf{h}_{u}^{(l)}) + \epsilon
\end{align}
where $\epsilon$ is a small constant introduced for numerical stability.
Eqn.~\ref{eqn:rho} means the rectified node features are directly used as neighbour messages.
For the message aggregation function $\boldsymbol{\zeta}^{(l)}(\cdot)$, we choose the form of softmax with a learnable temperature $\tau$:
\begin{align}
    \mathbf{m}_{v}^{(l)} 
    = \sum_{u \in \mathcal{N}(v)} \dfrac{e^{\frac{1}{\tau} \mathbf{m}_{vu}^{(l)}}}{\sum_{i \in \mathcal{N}(v)}e^{\frac{1}{\tau} \mathbf{m}_{vi}^{(l)}}} \cdot  \mathbf{m}_{vu}^{(l)}
\end{align}
which can be regarded as a weighted summation of all the neighbour messages.
The aggregated message $\mathbf{m}_{v}^{(l)}$ is then used to update the node feature of $v$ with a function $\boldsymbol{\phi}^{(l)}(\cdot)$:
\begin{align}
    \mathbf{h}_{v}^{(l+1)} = \boldsymbol{\phi}^{(l)}(\mathbf{h}_{v}^{(l)}, \mathbf{m}_{v}^{(l)})
    = \text{MLP}(\mathbf{h}_{v}^{(l)} + \mathbf{m}_{v}^{(l)})
\end{align}
where $\text{MLP}(\cdot)$ is a multi-layer perceptron with 2 fully-connected layers.

\begin{figure}[t]
	  \centering
      \subfigure[Visual branch]{
            \centering
	        \includegraphics[width=0.34\textwidth]{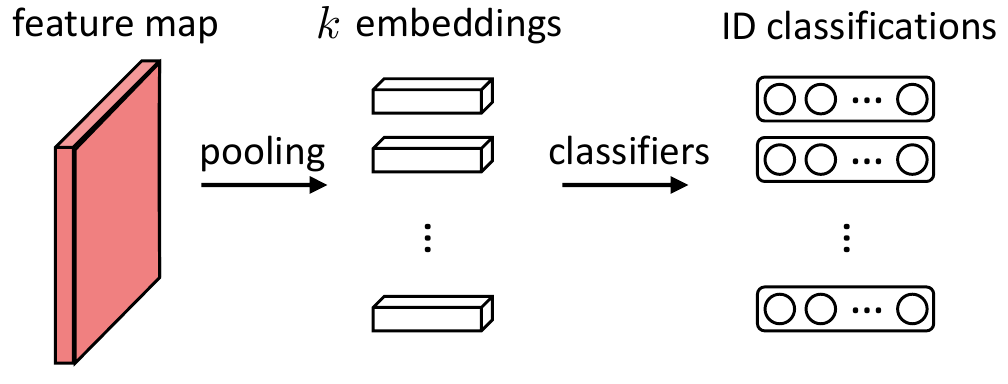}
            \label{fig:visual_pooling_supervision}}
      \subfigure[Graphical branch]{
            \centering
	        \includegraphics[width=0.45\textwidth]{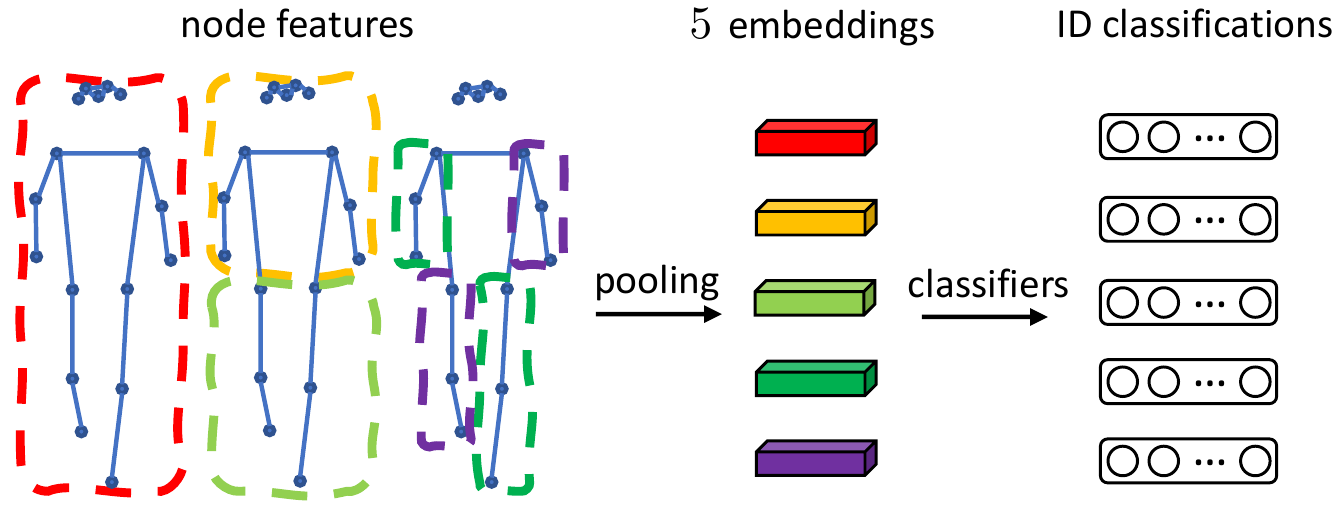}
            \label{fig:graphical_pooling_supervision}}
	  \caption{\textbf{Pooling and classification for visual and graphical branches.} (a) Spatial pooling on CNN feature maps. (b) Graph pooling~\cite{li2020model} on node features. Nodes wrapped with the same color are average pooled into a single embedding.}
	  \label{fig:pooling_supervision}
	  \vspace{-3pt}
\end{figure}

At the end of the GCN, the node features are pooled and mapped into discriminative embeddings.
An illustration is shown in Fig.~\ref{fig:graphical_pooling_supervision}.
We adopt the pooling operation in \cite{li2020model}, which combines multiple partition patterns and uses average pooling within each part.
Specifically, there are three types of discriminative embeddings, namely 1) average feature of the whole body, 2) the upper and lower body separately averaged features and 3) averaged pair features of (left arm, right leg) and (right arm, left leg).
All the features are then mapped with a fully connected layer without weight sharing.
In total, we obtain 5 discriminative embeddings, denoted as $\{\mathbf{x}_g^i | i=1,\ldots,5\}_t$ on frame $t$, which are further supervised by training objectives.

\subsection{Training \& Inference} 

For a video sequence with $T$ frames, a forward pass of our model generates two sets of embeddings from the visual branch and the graphical branch respectively.
The CNN embeddings from the visual branch is denoted as $\{\mathbf{x}_c^i | i=1,\ldots,k\}_t, t=1, \ldots, T$, where $k$ is a hyper-parameter depending on the CNN design.
For example, $k=1$ for an IDE model~\cite{zheng2017discriminatively} since global average pooling is applied on the final feature map, producing one single embedding for a frame.
For a PCB model~\cite{sun2018beyond}, $k$ equals the number of horizontal stripes which is typically set to $4$ or $6$.
Similarly, the output of the graphical branch is the GCN embeddings $\{\mathbf{x}_g^i | i=1,\ldots,5\}_t, t=1, \ldots, T$.
During training, all embeddings are input into $k + 5$ classifiers respectively.
Each classifier is composed of a fully connected layer and a softmax activation, which is supervised by a cross-entropy loss.

During inference, the embeddings on each frame are \emph{concatenated} and then \emph{averaged} over the temporal dimension, thus the CNN and GCN embeddings for the whole sequence is denoted as $\mathbf{x}_c$ and $\mathbf{x}_g$ respectively.
We find through experiments that $\mathbf{x}_c$ shows better performance than $\mathbf{x}_g$. 
Therefore, we report the re-ID results using only $\mathbf{x}_c$ unless mentioned otherwise.

\myparagraph{Discussion: Why discard GCN during inference?} 
Similar to dropping the discriminator in GANs and classifier layer in re-ID models, we also drop some part of our model during inference.
In our case, the entire graphical branch is dropped since we only use the CNN embedding $\mathbf{x}_c$.
The reason why we discard the graphical branch is twofold:
1) The forward pass of CNN does not rely on any information provided by the GCN.
2) The CNN embedding $\mathbf{x}_c$ performs better than GCN embedding $\mathbf{x}_g$.
It is worth to notice that although GCN is not used, it benefits CNN training via the back-propagated gradients.
In this way, $\mathbf{x}_c$ is enriched with long-range semantic dependencies and aligned temporal information, which is the fundamental advantage brought by keypoint message passing.

\begin{figure}[t]
    \centering
    \includegraphics[width=0.48\textwidth]{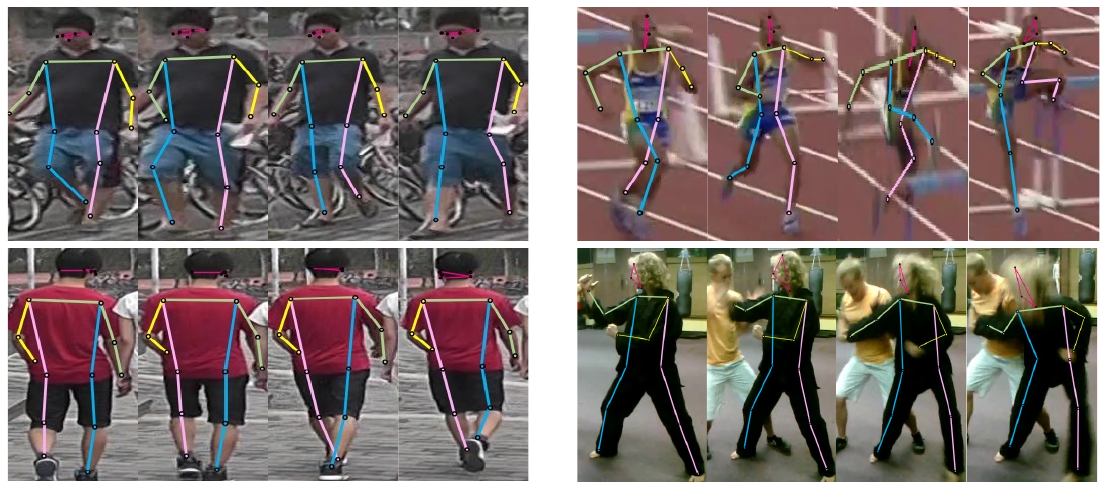}
    \caption{\textbf{Sample tracklets in MARS (left) and PoseTrackReID (right).} Person keypoints on MARS are generated by an off-the-shelf pose estimator~\cite{rafi2020selfsupervised}, whilst on PoseTrackReID the keypoints are manually annotated.}
    \label{fig:dataset_samples}
    \vspace{-3pt}
\end{figure}

\section{Experiments}
In this section, we first introduce the datasets, evaluation protocols and implementation details.
Then we compare our method to state-of-the-art methods, followed by extensive ablation studies.
More analytical experiments are included in our supplementary material.

\subsection{Datasets and Evaluation Protocol}
\textbf{MARS}~\cite{zheng2016mars} is a large-scale benchmark dataset for video-based person re-ID.
All the videos are collected with 6 stationary cameras on a university campus, from which a DPM detector~\cite{DPM} and GMMCP tracker~\cite{dehghan2015gmmcp} are used to crop out the person regions.
The training set consists of $8,298$ tracklets of $625$ identities, while the testing set includes $1,980$ tracklets of $626$ identities for query and $6,082$ tracklets of $620$ identities for gallery.
The person keypoints are generated with a top-down pose estimation model proposed in \cite{rafi2020selfsupervised} on each frame.
Some frames with visualized poses are shown in Fig.~\ref{fig:dataset_samples}.

\textbf{PoseTrackReID} is a new dataset proposed in this work to facilitate more comprehensive experiments for video-based person re-ID.
It is a cropped subset of the PoseTrack 2018 dataset~\cite{andriluka2018posetrack} which is originally proposed for multi-person pose estimation and articulated tracking.
The videos are captured in various scenes with large amounts of pose, appearance and scale variation. 
They also contain challenging scenes with severe body part occlusion and truncation.
The person keypoints are manually annotated at an interval of 4 frames at the beginning and at the end of a sequence, whereas the center of a sequence contains 30 consecutive manually annotated keypoints.
Based on PoseTrack 2018, we construct PoseTrackReID by adding additional annotations of person bounding box and global person ID.
The person regions are then cropped out for video-based person re-ID.
The training set of PoseTrackReID is gathered from the training set of PoseTrack 2018, including $7,725$ tracklets of $5,350$ identities.
The query set consists of $847$ tracklets of $830$ identities, while the gallery set includes $1,965$ tracklets of $1,696$ identities.
Both the query and gallery sets are collected from the validation set of PoseTrack 2018.
Some example frames are shown in Fig.~\ref{fig:dataset_samples}.

\myparagraph{Evaluation Protocols.} We use Cumulative Matching Characteristics~(CMC) and mean Average Precision~(mAP) as the evaluation metrics, which is the standard on the MARS benchmark. On PoseTrackReID, we follow the rules of MARS by using CMC and mAP as well.

\begin{table}[t]
  \centering
  \resizebox{1.0\columnwidth}{!}{
    \begin{tabular}{l|cc}
    \toprule
    \multicolumn{1}{c|}{\textbf{Method}} & \textbf{top-1} & \textbf{mAP} \\
    \midrule
    CNN + XQDA~\cite{zheng2016mars} & 65.3  & 47.6 \\ 
    SeeForest~\cite{zhou2017see} & 70.6  & 50.7 \\ 
    DuATM~\cite{si2018dual} & 81.2  & 67.7 \\ 
    Snippet~\cite{chen2018video}  & 86.3  & 76.1 \\ 
    ADFD~\cite{zhao2019attribute}  & 87.0 & 78.2 \\ 
    COSAM~(Subramaniam et al. 2019)  & 84.9  & 79.9 \\ 
    GLTR~\cite{li2019global}  & \textcolor[rgb]{0.23, 0.39, 0.70}{\textbf{87.0}} & 78.5 \\ 
    TCLNet~\cite{hou2020temporal}  & \textbf{88.8}  & \textcolor[rgb]{0.23, 0.39, 0.70}{\textbf{83.0}} \\ 
    KMPNet (ResNet-50)  & 86.7 & \textbf{84.4} \\
    \midrule
    ASTPN~\cite{xu2017jointly} & 44.0    & - \\ 
    VRSTC~\cite{hou2019vrstc}  & 88.5  & 82.3 \\ 
    MG-RAFA~\cite{zhang2020multi}  & 88.8  & 85.9 \\ 
    STGCN~\cite{yang2020spatial}  & \textcolor[rgb]{0.23, 0.39, 0.70}{\textbf{90.0}}    & 83.7 \\ 
    MGH~\cite{yan2020learning}   &  \textcolor[rgb]{0.23, 0.39, 0.70}{\textbf{90.0}}    & 85.8 \\ 
    KMPNet (PCB)  & 89.7  & \textcolor[rgb]{0.23, 0.39, 0.70}{\textbf{86.5}} \\
    KMPNet (MGH)  & \textbf{92.0}  & \textbf{86.6} \\
    \bottomrule
    \end{tabular}%
    }
  \caption{Performance comparison with state-of-the-art video-based person re-ID methods on MARS. ResNet-50, PCB and MGH in the brackets denote the base CNN of our visual branch. Methods in the upper block use global average pooling while the ones in the lower block use part-based pooling. Best performances in each block are marked \textbf{bold}, second best in \textcolor[rgb]{0.23, 0.39, 0.70}{\textbf{blue}}.}
  \label{tab:sota}%
  \vspace{-3pt}
\end{table}%

\subsection{Implementation}
We choose ResNet-50~\cite{He2016resnet} as the base CNN for the visual branch.
We adopt the 28-layer GCN model in \cite{li2020deepergcn} and remove the first graph convolution layer to match the visual branch.
We then partition the remaining 27 layers \textit{in proportion to} the design of ResNet-50. 
It is also straightforward to replace the ResNet-50 with other backbones.
The dimension for the latent node features of GCN is set to 64.
Please refer to the supplementary material for more details.

\subsection{Comparison to the state-of-the-art}
In this section, we compare our KMPNet to state-of-the-art methods on both MARS and PoseTrackReID.
We choose three representative base CNNs as the visual branch of our KMPNet, namely ResNet-50~\cite{He2016resnet}, PCB~\cite{sun2018beyond} and MGH~\cite{yan2020learning}.
The results on MARS are summarized in Tab.~\ref{tab:sota}.
For clear comparison, we group the methods in the table according to the pooling method used at the top convolutional layer. 
Methods in the upper block adopt global average pooling, while the lower block features part-based pooling.

Compared to the most recent methods, including MG-RAFA~\cite{zhang2020multi}, STGCN~\cite{yang2020spatial}, MGH~\cite{yan2020learning} and TCLNet~\cite{hou2020temporal}, our KMPNet with a simple PCB~\cite{sun2018beyond} as the visual branch achieves higher mAP and comparable top-1 accuracy.
Apart from the backbone CNN, all these methods require some extra computations such as spatial-temporal attention~\cite{zhang2020multi}, recursive feature erasing~\cite{hou2020temporal} and graph convolution~\cite{yang2020spatial,yan2020learning}.
In contrast, our method only requires graph convolution at the training stage.
During inference, no other computations are needed other than the backbone CNN, which makes our method computationally efficient.

Moreover, our method could also be used to boost the performance of other models by replacing the PCB baseline in the visual branch. 
For example, applying our keypoint message passing method to the MGH~\cite{yan2020learning} model\footnote{During training, the visual branch is initialized with the published model weight of MGH.} improves the top-1 accuracy and mAP by 2.0 and 0.8 pp. respectively, setting a new state-of-the-art performance.
This improvement also indicates that our method has good generalization ability \wrt different visual baselines. 

\begin{table}[t]
  \centering
    \begin{tabular}{l|cc} 
    \toprule
    \multicolumn{1}{c|}{\textbf{Method}} & \textbf{top-1} & \textbf{mAP} \\
    \midrule
    ResNet-50~\cite{He2016resnet} & 75.1  & 79.4 \\
    PCB~\cite{sun2018beyond}   & 77.9  & 81.5 \\
    MGH~\cite{yan2020learning}   & 82.7  & 84.2     \\ 
    \midrule 
    KMPNet (ResNet-50) &   78.7    &  82.7 \\
    KMPNet (PCB) & 79.2  & 82.7 \\ 
    KMPNet (MGH) & \textbf{83.3}  & \textbf{84.9}  \\
    \bottomrule 
    \end{tabular}%
    \caption{Performance comparison on PoseTrackReID.}
  \label{tab:sota_2}%
  \vspace{-3pt}
\end{table}%

We also show the re-ID results on PoseTrackReID in Tab.~\ref{tab:sota_2}.
The re-implemented CNNs are grouped in the upper block, while the corresponding GCN enhancements are listed in the lower block.
We can see from Tab.~\ref{tab:sota_2} that for all the three visual baselines, KMPNet consistently improves their top-1 accuracies and mAPs.
Compared to MARS, PoseTrackReID is more diverse with various background and human poses.
The fact that KMPNet also excels on PoseTrackReID suggests that our method is also generalizable to different scenes.

\subsection{Ablation Study}
In this subsection, we conduct analytical experiments on both MARS and PoseTrackReID.
For simplicity, we focus on one specific design which uses PCB~\cite{sun2018beyond} as the visual branch of our KMPNet.
The corresponding results and conclusions also apply to other base CNNs.

\begin{table}[t]
  \centering
    \begin{tabular}{l|cc|cc}
    \toprule
    \multicolumn{1}{c|}{\multirow{2}[4]{*}{\makecell{\textbf{Model} \\ \textbf{Variants}}}} & \multicolumn{2}{c|}{\textbf{MARS}} & \multicolumn{2}{c}{\textbf{PoseTrackReID}} \\
\cmidrule{2-5}          & \textbf{top-1} & \textbf{mAP}   & \textbf{top-1} & \textbf{mAP} \\
    \midrule
    PCB baseline & 85.3  & 84.6  & 77.9  & 81.5 \\
    + fine-tune & 84.7  &  84.6  & 77.2  & 81.4 \\
    \midrule
    + spatial &  88.6     &    85.5   & 78.9  & 82.1  \\
    + temporal &  88.7     &   85.3    & 77.9  & 81.8  \\
    + both &    89.7   &   86.5    & 79.2  & 82.7  \\
    \bottomrule
    \end{tabular}%
    \caption{Ablation study on spatial and temporal connections.}
  \label{tab:ablation}%
\end{table}%

\myparagraph{Analysis on the spatial temporal relationships.}
Compared to the PCB baseline, our model mainly benefits from two aspects, \ie the non-local spatial dependencies and the cross-frame temporal information.
In order to better understand the contribution of the two components, we conduct ablation studies on the spatial temporal structures of the graph.
Starting from the PCB baseline, which is basically the visual branch of our KMPNet, we add the graphical branch with different graph structures, namely \textit{spatial-only graph}, \textit{temporal-only graph} and \textit{spatial-temporal graph}. 
The three variants of graph structure are demonstrated in Fig.~\ref{fig:skelenton}.

The comparison is shown in the lower block of Tab.~\ref{tab:ablation}. 
We can see that adding spatial information with our method improves the performance of the PCB baseline by 3.3 and 0.9 pp. \wrt to top-1 and mAP on MARS,
which means that adding non-local information during training is beneficial for re-ID feature learning.
Meanwhile, the efficacy of temporal information is also clear: increasing top-1 and mAP by 3.4 and 0.7 pp.
Similarly, the experiment results on PoseTrackReID also reveal the same conclusion.
Finally, our final KMPNet featuring both spatial and temporal graph achieves the best performance over either of them alone. 

On the other hand, we also show in the upper block of Tab.~\ref{tab:ablation} a control experiment where the model is trained longer with the same learning rate and epochs as our KMPNet but without the assistance of the graphical branch. 
The re-ID accuracy of this model was not increased, which suggests that the performance gain is not due to the extra fine-tuning stage but the message passing via graphs.

Based on the above results, we could draw the conclusion that the strategy of guiding CNN training with spatial-temporal information and graph convolution is effective. 

\begin{table}[t]
  \centering
    \begin{tabular}{l|cc|cc}
    \toprule
    \multicolumn{1}{c|}{\multirow{2}[4]{*}{\textbf{Embedding}}} & \multicolumn{2}{c|}{\textbf{MARS}} & \multicolumn{2}{c}{\textbf{PoseTrackReID}} \\
\cmidrule{2-5}          & \textbf{top-1} & \textbf{mAP} & \textbf{top-1} & \textbf{mAP} \\
    \midrule
    $\mathbf{x}_c$    &  89.7     &  86.5     &   79.2    & 82.7  \\
    $\mathbf{x}_g$    &  61.7     &  57.3     &   48.3      & 51.6  \\
    \bottomrule
    \end{tabular}%
    \caption{Performance for visual/graphical embeddings.}
  \label{tab:embedding_ablation}%
  \vspace{-3pt}
\end{table}%

\myparagraph{Embedding choices.}
The two branches of our KMPNet produce two sets of embeddings respectively, \ie the CNN and GCN embeddings $\mathbf{x}_c$ and $\mathbf{x}_g$. 
In practice, we only use $\mathbf{x}_c$ for calculating the similarity between probe-gallery pairs.
What if we also take $\mathbf{x}_g$ into consideration? How would it affect the re-ID performance?
The answer to this question lies in Tab.~\ref{tab:embedding_ablation}, from which we can see that neither the top-1 accuracy nor the mAP of $\mathbf{x}_g$ are comparable to that of $\mathbf{x}_c$. 
The performance degeneration of $\mathbf{x}_g$ suggests that keypoint features processed by graph convolutions are not as expressive as CNN features, since they are just a sampled subset of CNN feature maps.
However, it does not obliterate the contribution of the GCN since it significantly boosts the performance of the CNN embedding $\mathbf{x}_c$, as is shown in Tab.~\ref{tab:ablation}.

Based on the analysis above, we decide to use only $\mathbf{x}_c$ for matching the query and gallery persons,
which makes it possible to remove the whole graphical branch during inference.
Therefore, the computation and memory resources needed are drastically reduced.

\myparagraph{Feature map visualizations.}
\begin{figure}[t]
    \centering
    \includegraphics[width=0.43\textwidth]{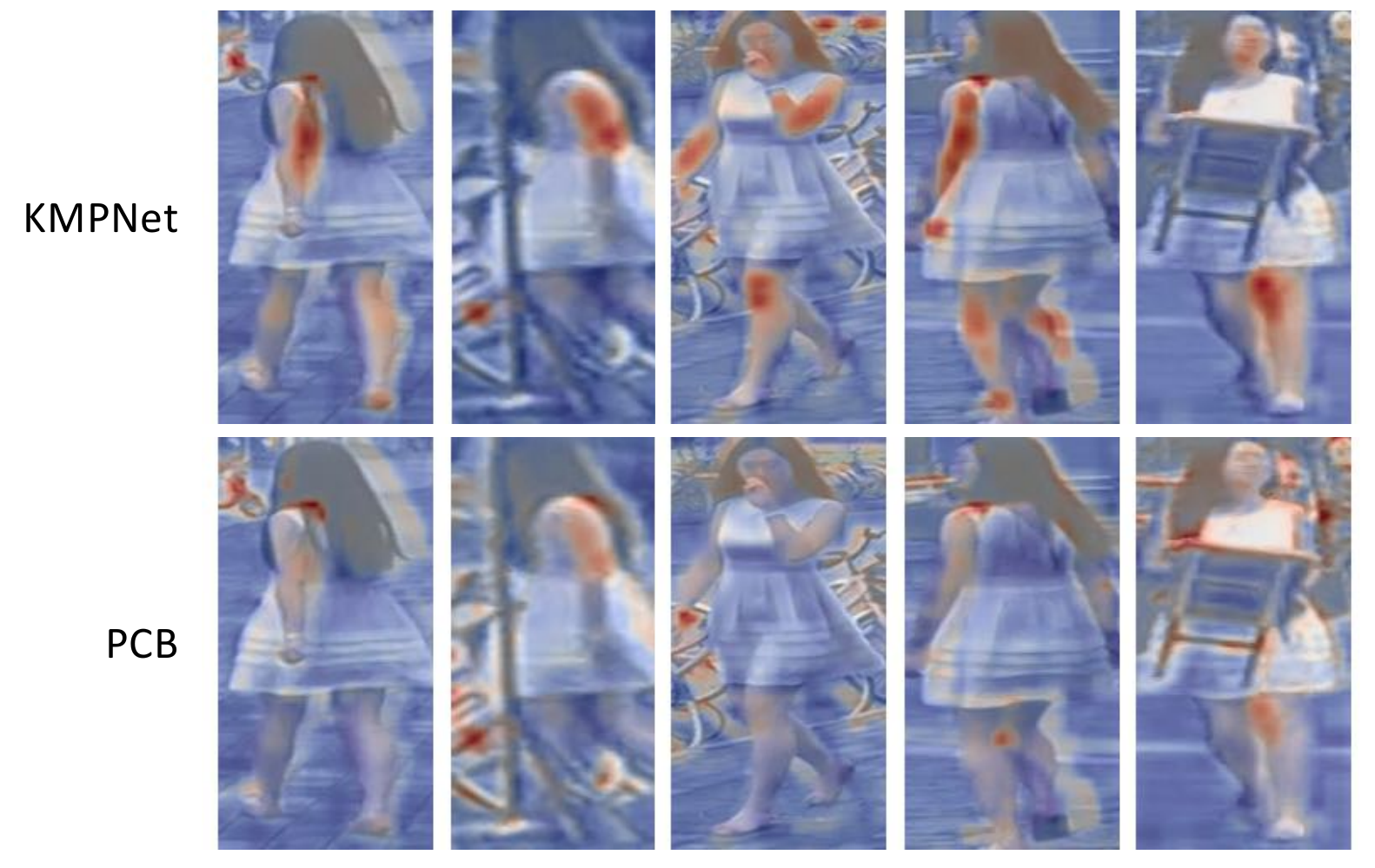}
    \caption{Feature map visualizations for our KMPNet and the PCB baseline. Warmer color denotes stronger activation.}
    \label{fig:vis}
    \vspace{-3pt}
\end{figure}
We visualize the feature maps of some representative samples in Fig.~\ref{fig:vis}.
The activation maps are obtained from the channel-wise max of features in `conv1'.
We can see from Fig.~\ref{fig:vis} that our KMPNet has stronger activation on the human body region than the baseline PCB model, despite that they have the same architecture during inference.
This is because the graphical branch in our KMPNet helps to cast stronger feedback signals onto the keypoint locations on CNN feature maps during training.
As a result, the learned visual branch tends to focus more on the human body region.
Therefore, more discriminative information could be discovered on human body.


\section{Conclusion}
In this paper, we present KMPNet, a spatial-temporal enhanced model for video-based person re-identification.
A graphical branch featuring a graph convolutional network is attached alongside a visual branch, which can be initialized with any CNN-based person re-ID model.
In the training stage, the graphical branch assists the CNN training by passing spatial and temporal messages on the feature maps, where spatial messages are passed among joint keypoints on the human body and temporal messages are passed between the same keypoints of adjacent video frames. 
During inference, the entire graphical branch can be dropped for efficiency, while the visual branch alone shows superior performance over the initial CNN model.
Extensive experiments on the MARS and PoseTrackReID dataset demonstrate the effectiveness of our method.


\section{Acknowledgement}
The authors would like to thank the AC and the anonymous reviewers for their critical and constructive comments and suggestions. 
This work has been funded by the National Science Fund of China (NSFC) (Grant No. U1713208 and 62172225),
Funds for International Cooperation and Exchange of NSFC (Grant No. 61861136011),
the Fundamental Research Funds for the Central Universities (No. 30920032201),
National Key R\&D Program of China (2017YFC0820601, 2021YFA1001100) 
and the Deutsche Forschungsgemeinschaft (DFG, German Research Foundation) - GA 1927/8-1. 
\bibliography{egbib}

\end{document}


\title{Keypoint Message Passing for Video-based Person Re-Identification \\
       Supplementary Material}

\author{
Supplementary Material
}

\maketitle

\section{Contents}
In this document, we provide additional materials to support our main submission. 
Specifically, we included the influence of kinematics feature to our keypoint message passing method in section \ref{sec:kin_supp}.

\section{The efficacy of kinematics feature.} \label{sec:kin_supp}
Our work is inspired by \cite{yan2018spatial} for action recognition and \cite{li2020model} for gait recognition, while both of them use keypoint coordinates as the node feature for input.
Their motivation behind this choice is to capture \textit{motion information}, which is vital for action and gait recognition.
In this section, we make a discussion on the efficacy of keypoint coordinates on our KMPNet for person re-identification.
In addition, we also investigate other motion-related features, such as the keypoint velocity and acceleration.
We collectively designate the three kind of features as \textit{kinematics feature}. 

\begin{figure}[t]
    \centering
    \includegraphics[width=0.45\textwidth]{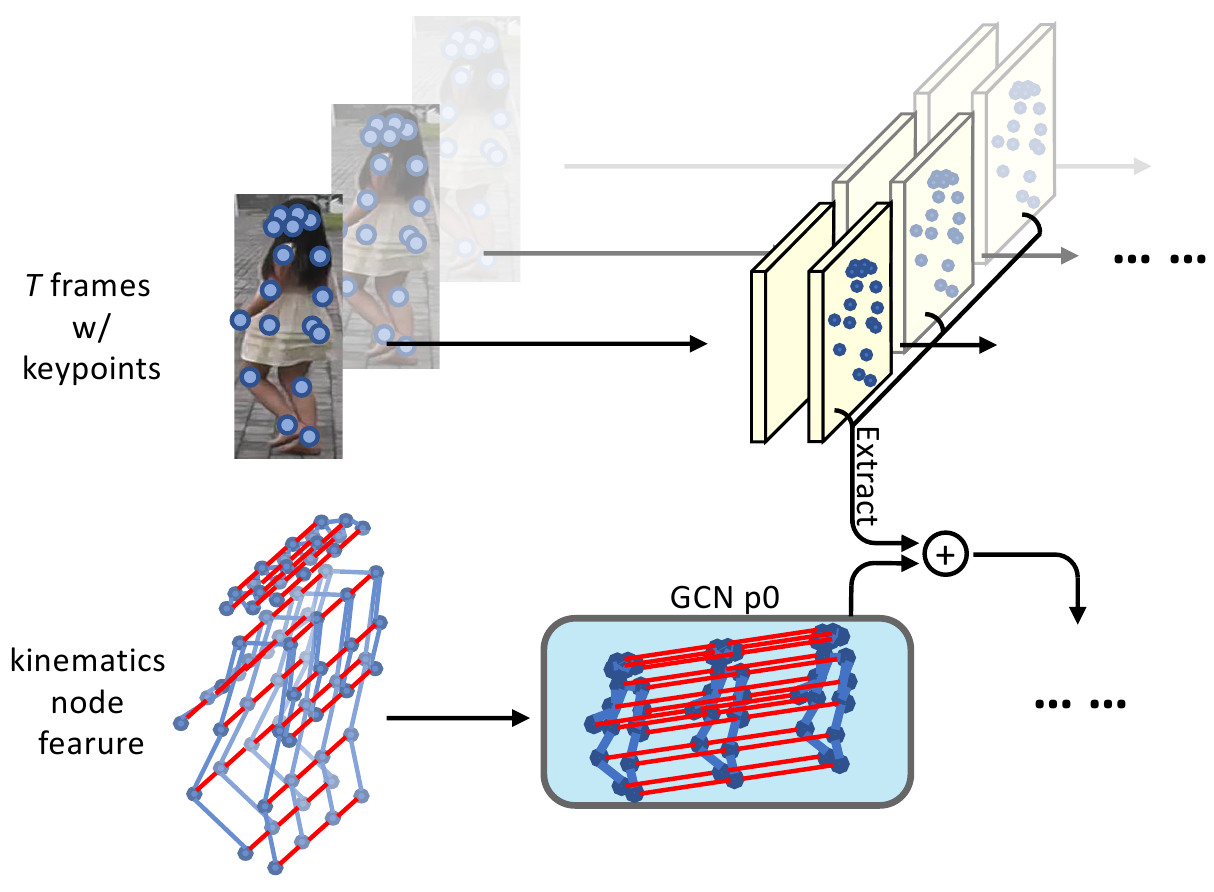}
    \caption{\textbf{Alternative design for our KMPNet with kinematics input.} The input node features are the combinations of keypoint coordinates, velocity and acceleration. The subsequent CNN and GCN layers are omitted for simplicity.}
    \label{fig:kinematics_input}
\end{figure}

In order to make use of kinematics features, we make a small modification to our KMPNet, which is shown in Fig.~\ref{fig:kinematics_input}.
We add a `GCN p0' at the very beginning of the graphical branch which takes in kinematics features as input.
Following \cite{li2020deepergcn}, `GCN p0' is composed of a fully-connected layer as node encoder and a graph convolutional layer.
The latent outputs of `GCN p0' are then fused with keypoint features extracted from CNN feature maps.
The rest part remains identical to our original KMPNet.

The performance of the modified KMPNet is shown in the upper block of Tab.~\ref{tab:kinematics_ablation}.
We can see that the performance is inferior to the proposed KMPNet without kinematic features, reducing the top-1 accuracy by 0.7 pp. with coordinate only on MARS and 0.2 pp. on PoseTrackReID.
Besides, using coordinate, velocity and acceleration at the same time even does more harm to the re-ID performance, \eg decreasing top-1 accuracy by 1.0 and 0.5 pp. on MARS and PoseTrackReID respectively.
Additionally, we also investigate the performance of a detached graphical branch as an individual GCN without the visual branch.
This individual GCN with only coordinate input yields very poor performance, achieving $2.3$ and $1.2$ pp. \wrt top-1 and mAP on MARS which is far behind the models with visual inputs.
We also get the same findings on PoseTrackReID.
This phenomenon suggests that motion information, which in this case is person gait, barely has any discrimination power for person re-identification. 
Our findings here are consistent with the results reported in \cite{zheng2016mars} \wrt motion features. 
Therefore, we draw the conclusion that kinematics features, including keypoint coordinates, velocity and acceleration, have a negative effect on our KMPNet, since re-ID focuses on the maker of the motion, rather than the motion itself.

\begin{table}[t]
  \centering
    \begin{tabular}{c|c|cc|cc}
    \toprule
    \multicolumn{2}{c|}{\textbf{Inputs}} & \multicolumn{2}{c|}{\textbf{MARS}} & \multicolumn{2}{c}{\textbf{PoseTrackReID}} \\
    \midrule
    \textit{visual} & \textit{graphical} & \textbf{top-1} & \textbf{mAP} & \textbf{top-1} & \textbf{mAP} \\
    \midrule
    video & coord.    & 89.0    & 86.1  &   79.0  &  82.6  \\
    video & (c, v, a) & 88.7  & 86.0    &  78.7     &  82.5 \\
    \midrule
    video &    -      & 89.7  & 86.5  & 79.2  & 82.7 \\
    \bottomrule
    \end{tabular}%
  \caption{\textbf{Efficacy discussion on the inputs to visual/graphical branches.} \textit{Upper block:} Alternative inputs of our KMPNet described in Fig~\ref{fig:kinematics_input}. 
  \textit{Lower block:} Our proposed KMPNet. `c', `v' and `a' are abbreviations for coordinate, velocity and acceleration respectively.}
  \label{tab:kinematics_ablation}%
\end{table}%

{\small
\bibliographystyle{ieee_fullname}
\bibliography{egbib}
}